\documentclass[journal=jpclcd,manuscript=letter,layout=twocolumn]{achemso}

\usepackage{chemformula} 
\usepackage[T1]{fontenc} 
\usepackage{algorithm}
\usepackage{algpseudocode}
\usepackage{amssymb}
\usepackage{booktabs}

\author{Yikai Liu}
\affiliation[Purdue ME]
{Department of Mechanical Engineering, Purdue University, West Lafayette, IN, 47906}
\email{liu3307@purdue.edu}
\author{Tushar K. Ghosh}
\affiliation[Purdue Chemistry]
{Department of Chemistry, Purdue University, West Lafayette, IN, 47906}
\email{tkghosh@purdue.edu}

\author{Guang Lin}
\affiliation[Purdue ME]
{Department of Mechanical Engineering, Purdue University, West Lafayette, IN, 47906}
\email{guanglin@purdue.edu}
\author{Ming Chen}
\email{chen4116@purdue.edu}
\affiliation[Purdue Chemistry]
{Department of Chemistry, Purdue University, West Lafayette, IN, 47906}

\title[An \textsf{achemso} demo]
  {Unbiasing Enhanced Sampling on a High-dimensional Free Energy Surface with Deep Generative Model}

\abbreviations{IR,NMR,UV}

\begin{document}


\begin{abstract}
Biased enhanced sampling methods utilizing collective variables (CVs) are powerful tools for 
sampling conformational ensembles. Due to high intrinsic dimensions, efficiently generating conformational ensembles for complex systems requires enhanced sampling on high-dimensional free energy surfaces. While methods like temperature-accelerated molecular dynamics (TAMD) can adopt many CVs in a simulation, unbiasing the simulation requires accurate modeling of a high-dimensional CV probability distribution, which is challenging for traditional density estimation techniques. Here we propose an unbiasing method based on the score-based diffusion model,
a deep generative learning method that excels in density estimation across complex data landscapes. We test the score-based diffusion unbiasing method on TAMD simulations. The results demonstrate that this unbiasing approach significantly outperforms traditional unbiasing methods, and can generate accurate unbiased conformational ensembles for simulations with a number of CVs higher than usual ranges. 
\begin{tocentry}
\includegraphics[width = 2.03in]{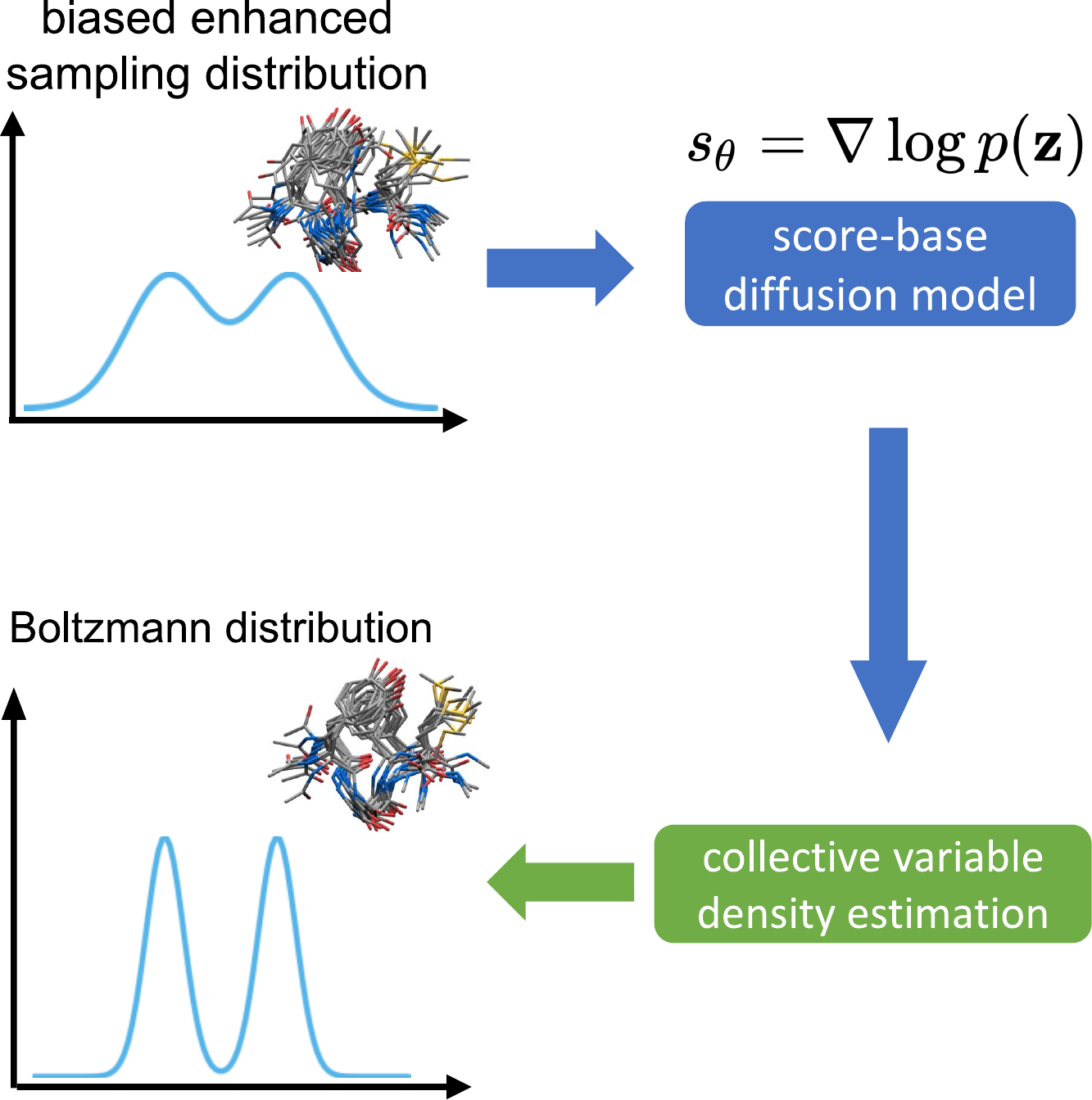}
\end{tocentry}
\end{abstract}

\par Molecular Dynamics (MD) simulations have emerged as a primary computational tool for studying the thermodynamic and kinetic properties of complex systems in chemistry, biology, and material science \cite{ciccotti2014molecular}. With the assistance of supercomputers \cite{shaw2014anton,shaw2021anton}, it is now possible to perform milliseconds of all-atom MD simulations for medium-sized proteins. 
To extend MD simulations to broader time and length scales, multiple enhanced sampling methods have been designed to 
increase MD sampling efficiency. \cite{laio2002escaping,barducci2008well,maragliano2006temperature,abrams2008efficient,darve2008adaptive,barducci2011metadynamics,abrams2010large,yu2011temperature,yu2014order,samanta2014microscopic,tiwary2013metadynamics}
Among all enhanced sampling methods, CV-based enhanced sampling methods focus on several important degrees of freedom that capture systems' essential dynamics. By biasing the probability distribution along CVs, CV-based enhanced sampling methods encourage systems to cross high energy barriers and explore different regions of the energy landscape more efficiently. 

\begin{figure}[h]
    \centering
    \includegraphics[width=3.33in]{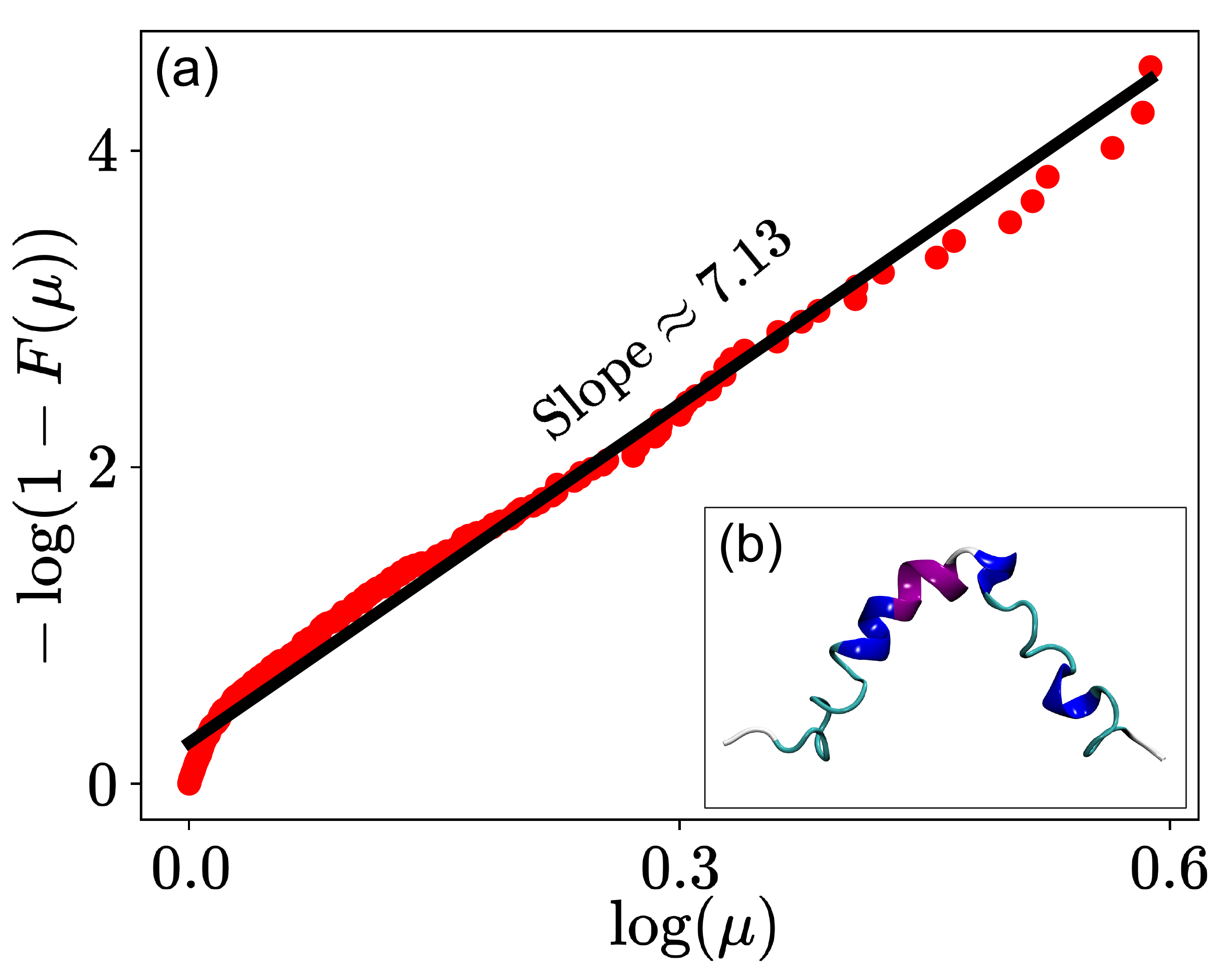}
    \caption{(a) 
    The Two Nearest Neighbors method calculates the ratio $\mu$ between the nearest and the second nearest 
    neighbour distance 
    for each data point (red dots). $F(\mu)$ is the accumulative distribution of $\mu$. The slope (black fitting line) of $-\log(1-F(\mu))$ as a function of $\log(\mu)$ 
    is an estimation of the intrinsic dimension of A$\beta_{42}$ (see the Supporting Information for details of the Two Nearest Neighbors method).
    (b) The molecular structure of A$\beta_{42}$.} 
    \label{fig:ID}
\end{figure}

A critical assumption behind CV-based enhanced sampling methods is the manifold hypothesis \cite{fefferman2016testing, narayanan2010sample,gorban2018blessing}, which posits that high-dimensional all-atom configurations often lie along a low-dimensional latent manifold, and such a low-dimensional manifold can accurately describe the important features of the high-dimensional systems. 
Traditionally, physics-based CVs are chosen from experimentally measurable properties, geometric descriptors and order parameters with important underlying physics, such as end-to-end distance \cite{hummer2001free}, radius of gyration \cite{bussi2006free}, or backbone torsion angles of proteins \cite{granata2013characterization}. Recently, machine-learning-based methods that utilize dimensionality reduction techniques have been applied to design CVs \cite{das2006low,ferguson2011integrating,rohrdanz2011determination,noe2015kinetic,zhang2018unfolding,fu2022simulate,ceriotti2011simplifying,tiwary2016spectral,mendels2018collective,ribeiro2018reweighted,wang2019past,nuske2014variational}. 
Regardless of CV categories, it is challenging to fully describe a complex system in biochemistry and material science with one or two parameters, a typical number of CVs used in many enhanced sampling simulations.
For example, following Two Nearest Neighbors method \cite{facco2017estimating}, we estimate the intrinsic dimension (minimum numbers of parameters) to accurately describe Amyloid-beta 42 (A$\beta_{42})$\cite{findeis2007role} to be 7.13, as shown in Fig.(\ref{fig:ID}). A number of CVs below 7 will not fully describe this system.

For most enhanced sampling methods \cite{chen2012heating,Parrinello2020gm,E2018reinforced,Gao2020vade} that utilize biasing potentials to assist crossing of energy barriers, it is challenging to use many CVs, primarily due to the difficulty of constructing accurate biasing potential for a high-dimensional FES. Other methods like driven-adiabatic free energy dynamics/temperature accelerated molecular dynamics (TAMD)\cite{maragliano2006temperature, abrams2008efficient, chen2012heating, cuendet2014free}, which enhance MD sampling by rising the temperature of certain degrees of freedom, are capable of adopting many CVs in enhanced sampling simulations, since biasing potential is not required.
However, when handling many CVs, the task of unbiasing TAMD trajectories to produce unbiased conformational ensembles becomes increasingly complex. This complexity arises from the need for accurate modeling of the high-dimensional, often multimodal CV probability distribution. Traditional density estimation techniques, such as histogram methods, kernel Density Estimation (KDE) \cite{chen2017tutorial}, Nearest Neighbor Density Estimation \cite{mack1979multivariate}, and Gaussian Mixture Model (GMM) \cite{reynolds2009gaussian}, often struggle to accurately capture the nuances of such intricate distributions.
For instance, KDE 
suffers in high-dimensional spaces as it may produce overly smooth or 
distorted estimates due to the lack of data across the expansive 
high-dimensional space. On the other hand, GMM suffers from scalability, initialization sensitivity, and a trivial model selection process. In this paper, we leverage the score-based diffusion model (SBDM)\cite{song2021score} for accurate unbiasing of enhanced sampling simulations with many CVs. 
In our study, we evaluate the performance of the SBDM-based unbiasing method in TAMD simulations. We will demonstrate that SBDM excels in constructing CV probability distributions, and can adapt to non-Euclidean CV such as torsion angles, with minor changes to the model architecture \cite{de2022riemannian}. These capacities endow the SBDM-based unbiasing method with superior performance and versatility.


\par We will first introduce the TAMD method and its unbiasing formula. 
For a system of $N$ particles, we denote its Cartesian coordinates by $\mathbf{r} \equiv (\mathbf{r}_1,\mathbf{r}_2,\ldots,\mathbf{r}_N)$, and 
$n$ collective variables by 
$\mathbf{q}\equiv (q_1(\mathbf{r}),\ldots,q_n(\mathbf{r}))$. In TAMD, $\mathbf{q}$ are 
coupled with extended variables $\mathbf{z}\equiv (z_1,\ldots,z_n)$ with stiff harmonic 
potentials $\sum_i\kappa_i/2(q_i(\mathbf{r})-z_i)^2$. $\mathbf{z}$ typically shares the same 
topology as $\mathbf{q}$. 
It has been proved that the free energy surface $A(\mathbf{q})$ can be approximated 
with the free energy surface of extended variables $A_{\kappa}(\mathbf{z})$ when 
$\kappa_i\rightarrow\infty$ for all $\kappa_i$. 
TAMD introduces a high temperature $T_{h} \gg T$ for $\mathbf{z}$, and maintains 
$\mathbf{r}$ at desired temperature $T$. To keep the thermodynamic properties of the system, 
$\mathbf{z}$ are adiabatically decoupled from $\mathbf{r}$
by assigning each $z_i$ a fictitious mass $\mu_i\gg 1$.
By defining $\beta_h=1/k_{\mathrm{B}}T_h$ where $k_{\mathrm{B}}$ is the Boltzmann constant, the joint probability distribution of $\mathbf{r}$ and $\mathbf{z}$ 
from a TAMD simulation satisfies \cite{chen2012heating}
\begin{equation}
    P_{\mathrm{TAMD}}(\mathbf{r},\mathbf{z}) \approx P_{T_h}(\mathbf{z})P(\mathbf{r}|\mathbf{z})\;\;,
    \label{eq:prob-tamd}
\end{equation}
where $P_{T_h}(\mathbf{z})\propto \exp\{-\beta_h A_{\kappa}(\mathbf{z})\}$ is 
the marginal probability distribution of $\mathbf{z}$ and $P(\mathbf{r}|\mathbf{z})$ is 
the Boltzmann distribution at the physical temperature $T$ conditional on $\mathbf{z}$. Eq.(\ref{eq:prob-tamd}) 
is exact if all $\mu_i\rightarrow\infty$. 
The free energy of collective variables $A_{\kappa}(\mathbf{z})$ can be easily obtained from TAMD with $A(\mathbf{z})\approx -k_{\mathrm{B}}T_h\log P_{T_h}(\mathbf{z})$.
\par We are often interested in intuitive 
\begin{figure}[h]
    \centering
    \includegraphics[width=3.33in]{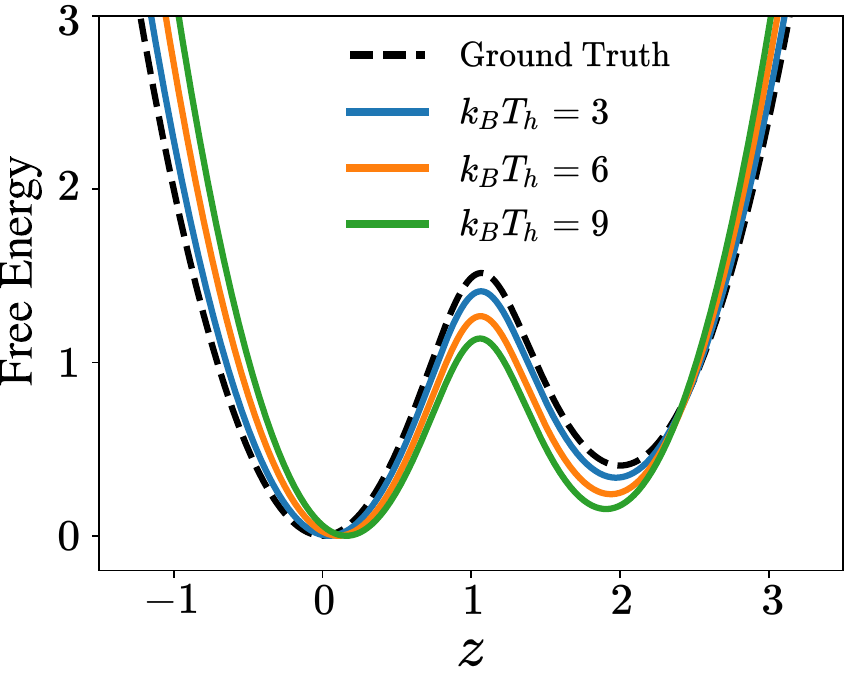}
    \caption{
    The free energy profiles generated from the unbiasing method of using Eq.(\ref{eq:unbiasing}) are shown as solid lines with an ``inaccurate'' estimation of 
    the high-temperature probability. The blue, orange, and green lines correspond to unbiasing results from $k_{\mathrm{B}}T_h=$3, 
    6, and 9, with $k_{\mathrm{B}}T=1$ in all three cases. The black dashed line is the ground truth with accurate estimation of the high-temperature probability. 
    }
    \label{fig:1D_Gaussian}
\end{figure}
features that are different from CVs used in an enhanced sampling 
to understand properties of the simulated system. Unbiasing enhanced sampling trajectories is necessary to 
project biased simulation data onto intuitive features. 
Assuming $\mathbf{Y}(\mathbf{r})$ is a set of low dimensional intuitive features of 
interests, the equilibrium probability of 
$\mathbf{Y}(\mathbf{r})=\mathbf{y}$ can be written as 
\begin{equation}
        P(\mathbf{y}) = \int \delta(\mathbf{y} - \mathbf{Y}(\mathbf{r}))\omega(\mathbf{z}) 
        P_{\mathrm{TAMD}}(\mathbf{r},\mathbf{z}) 
        \mathrm{d}\mathbf{r}\mathrm{d}\mathbf{z},
        \label{eq:unbiasing}
\end{equation}
where $\omega(\mathbf{z}) = P_{T_{h}}(\mathbf{z})^{T_{h}/T - 1}$ is the unbiasing weight. 
If $\mathbf{Y}$ can be written as a function of CVs, 
Eq.(\ref{eq:unbiasing}) can be reduced to $P(\mathbf{y}) = \int 
\delta(\mathbf{y} - \mathbf{Y}(\mathbf{z}))\omega(\mathbf{z})
P_{T_h}(\mathbf{z})\mathrm{d}\mathbf{z}$. 
\par A good estimation of $P_{T_{h}}(\mathbf{z})$ is crucial for obtaining an accurate $\omega(\mathbf{z})$ in TAMD. Errors in estimating $P_{T_{h}}(\mathbf{z})$ magnify errors in $\omega(\mathbf{z})$ at a high $T_h$, leading to 
inaccurate $P(\mathbf{y})$. We design a toy problem to demonstrate the importance of accurately modeling 
$P_{T_{h}}(\mathbf{z})$ in unbiasing TAMD. In this example, a one dimensional probability density $P(\mathbf{z})$ at 
$k_{\mathrm{B}}T=1$ is constructed with a mixture of 
Gaussians. The probability is 
\begin{figure*}[h]
    \centering
    \includegraphics[width=5in]{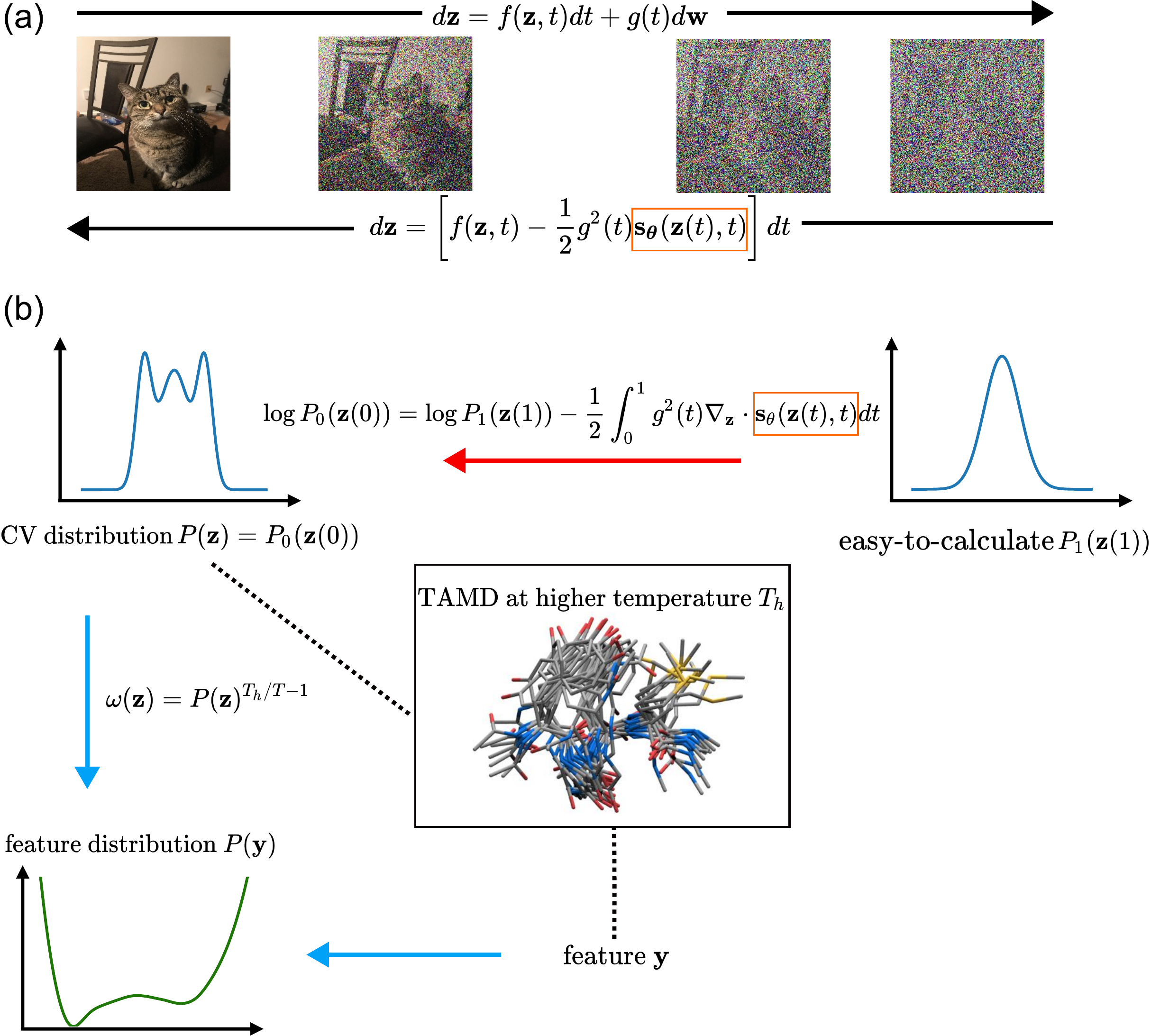}
    \caption{(a) The mechanism of denoising SBDM. During the diffusion process, the data (in this demonstration a picture) is gradually perturbed to an isotropic Gaussian noise via a forward SDE. 
    (b) The flow chat of unbiasing TAMD with SBDM. We perform TAMD simulation at a high temperature $T_h$, and construct SBDM for $\mathbf{z}$. The time-dependent score function, squared in orange, is used to perform density estimation of CV, $P(\mathbf{z})\equiv P_{T_h}(\mathbf{z})$, from a simple distribution, as shown above the red line. The weight $\omega(\mathbf{z})$ of configurations is evaluated from the estimated CV probability $P(\mathbf{z})$. The weight is used to compute the unbiased distribution of features of interest $P(\mathbf{y})$.}
    \label{fig:score_frame}
\end{figure*}
scaled to different $T_h$ followed by convolution with a Gaussian kernel to ``perturb'' 
the high temperature probability $\tilde{P}_{T_{h}}(\mathbf{z})$. 
$\tilde{P}_{T_{h}}(\mathbf{z})$ is used to calculate weight $\omega(\mathbf{z})$. The ``unbiased'' distribution of $\mathbf{z}$, $P_{\mathrm{ub}}(\mathbf{z})$, is then calculated with Eq.(\ref{eq:unbiasing}) (see the Supporting Information for details of the toy problem). 
Fig.(\ref{fig:1D_Gaussian}) demonstrates that the error in $P_{\mathrm{ub}}(\mathbf{z})$ is more sensitive to errors in 
$\tilde{P}_{T_{h}}(\mathbf{z})$ when $T_h$ is higher. 

\par We then briefly review the SBDM framework and its usage in density estimation. The SBDM perturbs data to noise prior with a diffusion process over a unit time by a linear stochastic differential equation (SDE):
\begin{equation}
    d \mathbf{z}=f(\mathbf{z}, t) d t+g(t) d \mathbf{w}, \; t\in [0,1],
    \label{Foward_sde}
\end{equation}
where $f(\mathbf{z},t), g(t)$ are user-defined drift and diffusion functions of the SDE and $\mathbf{w}$ denotes a standard Wiener process. In this paper, an SDE with the drift term $f(\mathbf{z},t) = 0$ is used. 
With carefully designed $g(t)$, the marginal probability of 
$\mathbf{z}$ at diffusion time $t$, $P_t(z)$, changes 
from the data distribution at $t=0$ to approximately a simple Gaussian distribution at $t=1$. 
\par For any diffusion process in Eq.(\ref{Foward_sde}), it has a corresponding reverse-time SDE \cite{anderson1982reverse} :
\begin{equation}
    d \mathbf{z} = [f(\mathbf{z},t) - g^{2}(t)\nabla_{\mathbf{z}}\log P_{t}(\mathbf{z})]dt +g(t)d\bar{\mathbf{w}},
    \label{rev_SDE}
\end{equation}
with $\bar{\mathbf{w}}$ a standard Wiener process in the reverse-time. The trajectories of the reverse SDE have the same marginal densities as the forward SDE. Thus, the reverse-time SDE can gradually convert noise to data. The SBDM parameterizes the time-dependent score function $\nabla_{\mathbf{z}}\log P_{t}(\mathbf{z})$ in the reverse SDE with a neural network $\mathbf{s}_{\boldsymbol{\theta}}(\mathbf{z}(t), t)$. To estimate $\nabla_{\mathbf{z}}\log P_{t}(\mathbf{z})$, a time-dependent score-based model $\mathbf{s}_{\boldsymbol{\theta}}(\mathbf{z}(t), t)$ can be trained via minimizing a denoising score matching loss: 
\begin{equation}
\begin{aligned}
M(\theta)&=\left\|\mathbf{s}_{\boldsymbol{\theta}}(\mathbf{z}(t), t)-\nabla_{\mathbf{z}(t)} \log P(\mathbf{z}(t)|\mathbf{z}(0))\right\|_2^2 \\
J(\theta)&=\underset{\boldsymbol{\theta}}{\arg \min } \mathbb{E}_t\left\{ \mathbb{E}_{\mathbf{z}(0)} \mathbb{E}_{\mathbf{z}(t) \mid \mathbf{z}(0)}M(\theta)\right\},
\end{aligned}
\label{sde_loss}
\end{equation}
with $t$ uniformly sampled between $[0,1]$. We note 
that $\nabla_{\mathbf{z}(t)} \log P(\mathbf{z}(t)|\mathbf{z}(0))$ is not explicitly required in the score matching loss 
while samples following 
$P(\mathbf{z}(t)|\mathbf{z}(0))$ are needed. 

Finally, SBDM defines a deterministic way to compute the data distribution $P_0(\mathbf{z})$ as follows, with $f(\mathbf{z},t)=0$:
\begin{align}
   \log P_{0}(\mathbf{z}(0)) & = \log P_{1}(\mathbf{z}(1)) \nonumber \\
   &  - \frac{1}{2}\int_{0}^{1} g^{2}(t)\nabla_{\mathbf{z}}\cdot \mathbf{s}(\mathbf{z}(t),t)dt
    \label{density}.
\end{align}
\par CVs can be defined in spaces with different topologies. 
For example, $n$ torsion angles are defined on a hypertorus space $\mathbb{T}^{n}$, and quaternions which representing 
rigid-body rotation are defined on a 3D unit sphere $\mathbb{S}^{3}$. Therefore, extending the SBDM to different topologies is important. In this study we will focus on SBDM on a hypertorus space.  The theory behind SBDM holds for compact Riemannian manifolds, with subtle modifications. For $\mathbf{z} \in M$, a Riemannian manifold (such as hypertorus $\mathbb{T}^{n}$), $\mathbf{w}$ being the Brownian motion on the manifold, and $f(\mathbf{z},t) \in T_{\mathbf{z}}M$, a tangent space, Eq.(\ref{rev_SDE}) still holds \cite{de2022riemannian}.
 \par As shown in Eq.(\ref{sde_loss}), training a denoising score matching model requires sampling from the perturbation kernel $P(\mathbf{z}(t)|\mathbf{z}(0))$ of the forward diffusion defined by Eq.(\ref{Foward_sde}). We consider the perturbation kernel on $\mathbb{T}^{n}$ with wrapped normal distribution:
\begin{equation}
\begin{aligned}
  & \mathbf{U}_{\mathbf{d}}(t) = \frac{\left\|\boldsymbol{\mathbf{z}}(0)-\boldsymbol{\mathbf{z}}(t)+2 \pi  \mathbf{d}\right\|^2}{2 \sigma^2(t)}, \; \mathbf{d} \in \mathbb{Z}^n\\
  & P(\mathbf{z}(t)|\mathbf{z}(0)) \propto \sum_{\mathbf{d} } \exp \left(-\mathbf{U}_{\mathbf{d}}(t)\right),
\end{aligned}
\end{equation} where $g(t)$ and $\sigma(t)$ are related by $g(t)=\sqrt{\mathrm{d}\sigma^2(t)/\mathrm{d}t}$. The rest of the terms in the loss function in Eq. (\ref{sde_loss}) remain the same. 
\begin{figure}[h]
    \centering
    \includegraphics[width=3.33in]{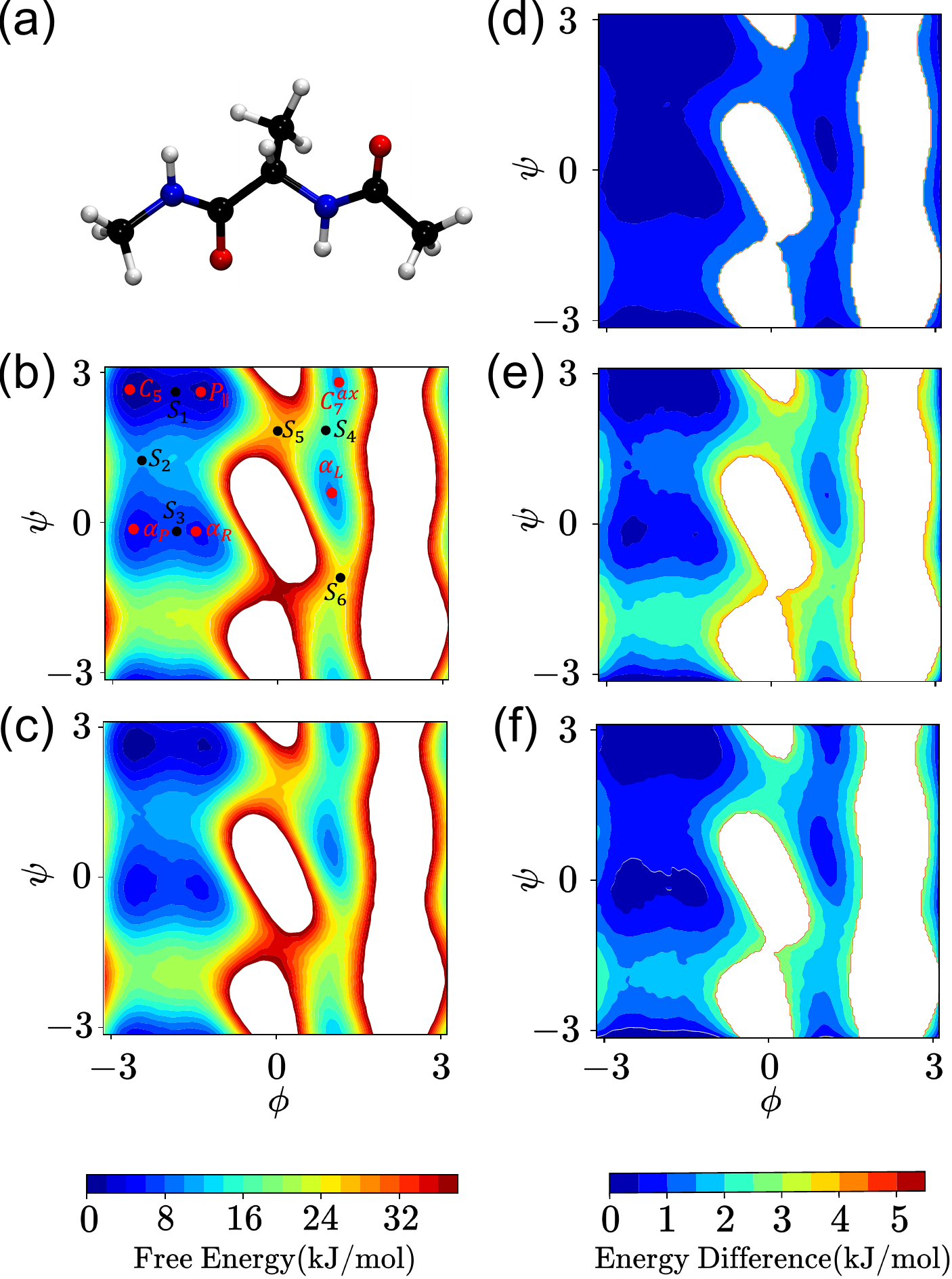}
    \caption{Molecular structure of alanine dipeptide is shown in (a). (b, c) FES w.r.t backbone dihedral angles of alanine dipeptide are presented, obtained by metadynamics (b) and SBDM-TAMD (c). The red/black points represent the location of important minima and saddle points on the FES. (d-f) The absolute free energy difference between metadynamics and SBDM-TAMD, KDE-TAMD, and GMM-TAMD.}
    \label{fig:ala}
\end{figure} Note that density estimation in Eq.(\ref{density}) is applicable to both variables in Euclidean and hypertorus space.
The architecture of an SBDM model 
is highly flexible. For example, an SBDM model can use Residual neural network (ResNet) 
\cite{he2016deep}, U-Net \cite{olaf2015unet}, Graph Neural Network (GNN) \cite{scarselli2008graph}, etc. 

\par In the following section, we will demonstrate how incorporating SBDM can fulfill the strict density estimation accuracy requirement of unbiasing TAMD, thus allowing TAMD to generate correct unbiased ensembles. We tested the efficiency and accuracy of SBDM to unbiasing TAMD simulations on three systems: 
(1) alanine dipeptide (2) glutamine dipeptide, and (3) met-enkephalin 
(see the Supporting Information for simulation details).  
In all three systems, we conducted TAMD with torsion angles as collective variables, with $T_{h} = 1200 K$ in the first example while $T_{h} = 900 K$ in the second and third examples. The physical variables were maintained at temperature $T = 300 K$ in all three experiments. We then unbiased TAMD with density estimation performed by SBDM on hypertorus space (see the Supporting Information for details of training SBDM models). 
Unbiasing TAMD with SBDM as density estimation method is referred as SBDM-TAMD in the rest of the paper. The general framework of SBDM-TAMD is summarized in Fig.(\ref{fig:score_frame}). For a fair comparison, we unbiased the same TAMD simulations with kernel density estimation (KDE-TAMD) and Gaussian Mixture Model (GMM-TAMD) in all three systems, and Normalizing Flow on hypertorus space (NF-TAMD) \cite{rezende2020normalizing,kohler2021smooth} in the last system. We performed converged metadynamics simulations at $300K$ as the baseline results of all three systems. 
We will compare the unbiasing results from different density estimation methods to the baseline results. 

\par The first proof-of-concept example is an alanine dipeptide in the aqueous solution with implicit solvent. 
This has been a benchmark system with well-established FES in previous studies. \cite{strodel2008free,smith1999alanine} In both TAMD and the baseline metadynamics, we used the Ramachandran angles ($\phi$, $\psi$) as CVs. The accuracy of SBDM-TAMD is demonstrated in 
Fig.(\ref{fig:ala}). The SBDM capably captures the intricate free energy delineated by metadynamics, as shown in Fig.(\ref{fig:ala})(c) and Fig.(\ref{fig:ala})(d). Notably, the model accurately locates all six free energy minima, with depth of the minima quantitatively accurate (<1 kJ/mol energy difference). Furthermore, the saddle points - representative of transitional between conformational basins, are also well reproduced. 
Energy errors in low free energy saddle points are less than 2 kJ/mol compared to metadynamics results. 
This match implies that the SBDM-TAMD is accurate for studying 
\begin{figure}[h]
    \centering
    \includegraphics[width=3.33in]{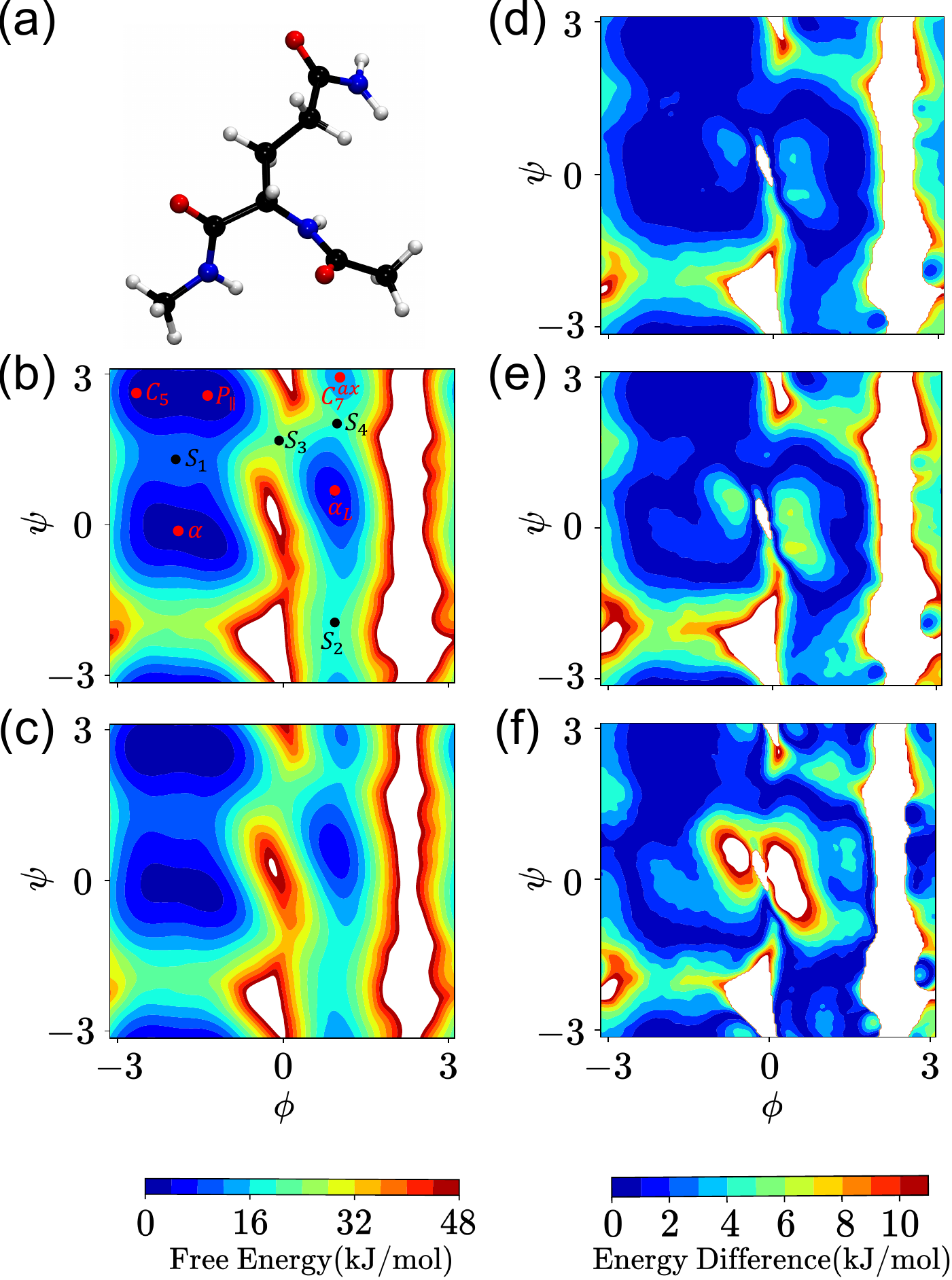}
    \caption{Molecular structure of Glutamine dipeptide is shown in (a). FES of backbone dihedral angles of Glutamine dipeptide are calculated by metadnyamics (b) and SBDM-TAMD (C). The red/black points represent the location of important minima and saddle points on the FES. (d)-(f) shows he absolute free energy difference between metadynamics and SBDM-TAMD,  KDE-TAMD, and GMM-TAMD.}
    \label{fig:Q1}
\end{figure}
both thermodyanmics and kinetics of alanine dipeptide conformational 
changes. 
We notice that for this two-dimensional problem, KDE and GMM perform relatively well, but still with less accuracy compared to SBDM. Both methods can locate the free energy minima but with a larger energy difference of up to 2.5 kJ/mol from the baseline metadynamics. The free energy of saddle points have even larger deviation, up to 3.5 kJ/mol. The clear disparities in free energy differences, as shown in Fig. \ref{fig:ala}(e) and Fig. \ref{fig:ala}(f), especially in regions corresponding to the annotated energy minima and saddle points, illustrate the inadequacies of KDE and GMM for precise FES calculations.

The second system we studied is Glutamine dipeptide in the aqueous solution with explicit solvent. A TAMD simulation was performed with five dihedral angles on the backbone and the side chain ($\phi, \psi, \chi_1, \chi_2, \chi_3$) as CVs to enhance the sampling of 
both backbone 
and side chain conformations. We also run a benchmark metadynamics with two backbone diheral angles as CVs. The result is demonstrated in Fig.(\ref{fig:Q1}). Projecting unbiased SBDM-TAMD trajectory onto backbone dihedral angles quantitavely matches the bechmark FES from 
metadynamics. Energy errors of minima on the projected 2D FES are within 2 kJ/mol while errors of low-free-energy saddles are 
approximately 5 kJ/mol. 
As a comparison, both KDE and GMM struggle to uphold their precision with the increased CV dimensions. Both of these two methods 
introduce larger errors at $\alpha$ and $\alpha_L$. 
This example 
highlights the SBDM's superior adaptability and accuracy in high-dimensional CV compared to traditional methods.
\par
\begin{figure}[h]
    \centering
    \includegraphics[width=3.33in]{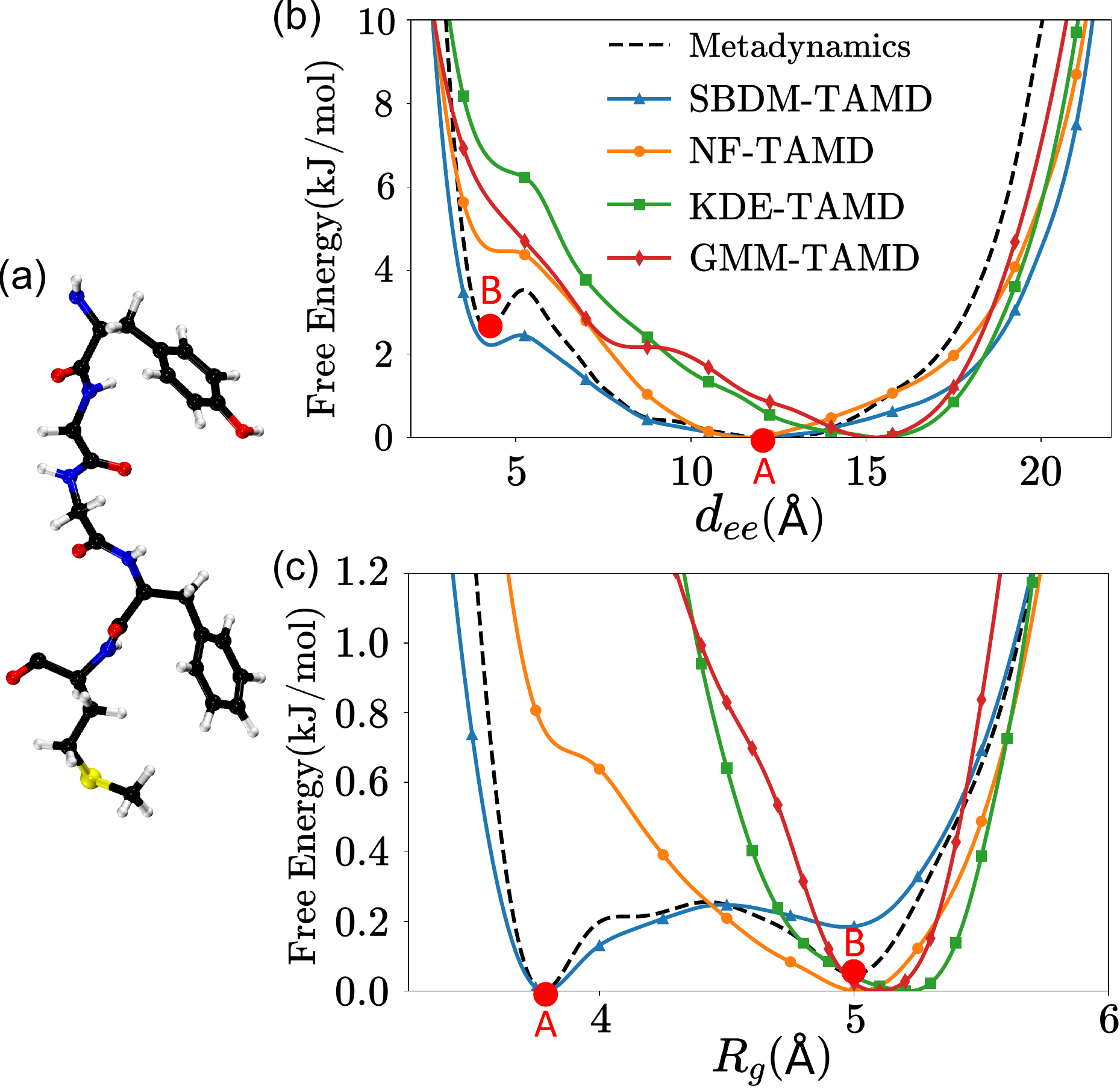}
    \caption{(a) Molecular structure of met-enkephalin. (b-c) Free energy of end-to-end distance and radius of gyration, with red dots indicating stable and meta-stable states.}
    \label{fig:3}
\end{figure}

 The final, more challenging system we studied is the oligopeptide met-enkephalin (Tyr-Gly-Gly-Phe-Met) in the aqueous solution with explicit water, which is a common test case for enhanced sampling techniques. \cite{Klein2010tdbias,chen2012heating,chen2015locating} For TAMD, we chose ten backbone dihedral angles ($\phi_1, \psi_1,...,\phi_5, \psi_5$) as CVs. The baseline metadynamics simulation was performed with a 2D stochastic kinetic embedding (StKE),  a manifold learning method which serves as a low-dimensional CV representation that preserves kinetic information \cite{zhang2018unfolding}. 
We projected the unbiased SBDM-TAMD trajectory to features like end-to-end distance $d_{\mathrm{ee}}$ and the radius of gyration $R_{\mathrm{g}}$. 
As a comparison, we unbiased the benchmark metadynamics simulation \cite{sutto2010comparing,sutto2012new,sicard2013reconstructing} 
and projected configurations from metadynamics onto the same features. We want to emphasize that generating an optimal machine-learning-based 
CVs for metadynamics (StKE in this work) is non-trivial and requires extra simulation data. 
From Fig.(\ref{fig:3}), the SBDM-TAMD exhibits high performance, demonstrating a FES that closely aligned with the metadynamics baseline. The minima and barriers on each projected one-dimensional FES predicted by SBDM is in 
good agreements with benchmark, suggesting SBDM-TAMD 
is capable of modeling thermodynamics and kinetics of a polypeptide. 
For example, as shown in Fig. \ref{fig:3}(b), both metadynamics and SBDM-TAMD identify conformations with small $d_{\mathrm{ee}}$ 
as a meta-stable conformation with consistent meta-stability, while KDE and GMM fail to predict this meta-stable conformation. 
Also, KDE and GMM fail to predict the location of the global minimum on the FES of $d_{\mathrm{ee}}$ with an error of $\sim$3.0 \AA. Although NF correctly predicts 
the location of the global minimum, it overestimates the stability of this minimum. 
Similarly, KDE ,GMM, and even NF, 
predict the most possible $R_{\mathrm{g}}\approx 5$ \AA, while 
SBDM-TAMD predicts that the most stable 
conformations have $R_{\mathrm{g}}<4$ \AA, which is consistent with the metadynamics' 
result.
\par In this paper, we developed an unbiasing method based on score-based diffusion model, a deep generative learning model, to generate unbiased conformational ensembles from collective-variable-based biased enhanced sampling simulations. Our method can adapt to simulations with collective variables of large amounts and different topologies. We test the unbiasing method on temperature-accelerated molecular dynamics, an enhanced sampling method that can utilize many collective variables to efficiently explore a high-dimensional free energy surface. Numerical experiments across three systems of 2,5 and 10 collective variables underscore our unbiasing method's exceptional accuracy and adaptability to high-dimensional collective variables. Although biomolecules were used in the numerical 
experiments, the developed approach can be used to other problems like 
modeling material phase transitions. 
Looking ahead, we aim to explore the method's potential in unbiasing simulations with even larger amounts of collective variables, and form accurate high-dimensional biasing potentials for wider ranges of enhanced sampling methods. 
\begin{acknowledgement}
Y.L., T.K.G. and M.C. acknowledge support from Purdue startup funding. G.L. and Y.L. gratefully acknowledge the support of the National Science Foundation (DMS-2053746, DMS-2134209, ECCS-2328241, and OAC-2311848), and U.S. Department of Energy (DOE) Office of Science Advanced Scientific Computing Research program DE-SC0021142, DE-SC0023161, and the Uncertainty Quantification for Multifidelity Operator Learning (MOLUcQ) project (Project No. 81739), and DOE – Fusion Energy Science,  under grant number: DE-SC0024583.  
\end{acknowledgement}

\begin{suppinfo}
Supporting information contains details of MD simulations, 
SBDM training, Two Nearest Neighbors intrinsic dimension estimation, 
and toy model.
\end{suppinfo}


\bibliography{achemso-demo}

\providecommand{\latin}[1]{#1}
\makeatletter
\providecommand{\doi}
  {\begingroup\let\do\@makeother\dospecials
  \catcode`\{=1 \catcode`\}=2 \doi@aux}
\providecommand{\doi@aux}[1]{\endgroup\texttt{#1}}
\makeatother
\providecommand*\mcitethebibliography{\thebibliography}
\csname @ifundefined\endcsname{endmcitethebibliography}
  {\let\endmcitethebibliography\endthebibliography}{}
\begin{mcitethebibliography}{58}
\providecommand*\natexlab[1]{#1}
\providecommand*\mciteSetBstSublistMode[1]{}
\providecommand*\mciteSetBstMaxWidthForm[2]{}
\providecommand*\mciteBstWouldAddEndPuncttrue
  {\def\EndOfBibitem{\unskip.}}
\providecommand*\mciteBstWouldAddEndPunctfalse
  {\let\EndOfBibitem\relax}
\providecommand*\mciteSetBstMidEndSepPunct[3]{}
\providecommand*\mciteSetBstSublistLabelBeginEnd[3]{}
\providecommand*\EndOfBibitem{}
\mciteSetBstSublistMode{f}
\mciteSetBstMaxWidthForm{subitem}{(\alph{mcitesubitemcount})}
\mciteSetBstSublistLabelBeginEnd
  {\mcitemaxwidthsubitemform\space}
  {\relax}
  {\relax}

\bibitem[Ciccotti \latin{et~al.}(2014)Ciccotti, Ferrario, Schuette,
  \latin{et~al.} others]{ciccotti2014molecular}
Ciccotti,~G.; Ferrario,~M.; Schuette,~C., \latin{et~al.}  Molecular dynamics
  simulation. \emph{Entropy} \textbf{2014}, \emph{16}, 1\relax
\mciteBstWouldAddEndPuncttrue
\mciteSetBstMidEndSepPunct{\mcitedefaultmidpunct}
{\mcitedefaultendpunct}{\mcitedefaultseppunct}\relax
\EndOfBibitem
\bibitem[Shaw \latin{et~al.}(2014)Shaw, Grossman, Bank, Batson, Butts, Chao,
  Deneroff, Dror, Even, Fenton, \latin{et~al.} others]{shaw2014anton}
Shaw,~D.~E.; Grossman,~J.; Bank,~J.~A.; Batson,~B.; Butts,~J.~A.; Chao,~J.~C.;
  Deneroff,~M.~M.; Dror,~R.~O.; Even,~A.; Fenton,~C.~H. \latin{et~al.}  Anton
  2: raising the bar for performance and programmability in a special-purpose
  molecular dynamics supercomputer. SC'14: Proceedings of the International
  Conference for High Performance Computing, Networking, Storage and Analysis.
  2014; pp 41--53\relax
\mciteBstWouldAddEndPuncttrue
\mciteSetBstMidEndSepPunct{\mcitedefaultmidpunct}
{\mcitedefaultendpunct}{\mcitedefaultseppunct}\relax
\EndOfBibitem
\bibitem[Shaw \latin{et~al.}(2021)Shaw, Adams, Azaria, Bank, Batson, Bell,
  Bergdorf, Bhatt, Butts, Correia, \latin{et~al.} others]{shaw2021anton}
Shaw,~D.~E.; Adams,~P.~J.; Azaria,~A.; Bank,~J.~A.; Batson,~B.; Bell,~A.;
  Bergdorf,~M.; Bhatt,~J.; Butts,~J.~A.; Correia,~T. \latin{et~al.}  Anton 3:
  twenty microseconds of molecular dynamics simulation before lunch.
  Proceedings of the International Conference for High Performance Computing,
  Networking, Storage and Analysis. 2021; pp 1--11\relax
\mciteBstWouldAddEndPuncttrue
\mciteSetBstMidEndSepPunct{\mcitedefaultmidpunct}
{\mcitedefaultendpunct}{\mcitedefaultseppunct}\relax
\EndOfBibitem
\bibitem[Laio and Parrinello(2002)Laio, and Parrinello]{laio2002escaping}
Laio,~A.; Parrinello,~M. Escaping free-energy minima. \emph{Proceedings of the
  national academy of sciences} \textbf{2002}, \emph{99}, 12562--12566\relax
\mciteBstWouldAddEndPuncttrue
\mciteSetBstMidEndSepPunct{\mcitedefaultmidpunct}
{\mcitedefaultendpunct}{\mcitedefaultseppunct}\relax
\EndOfBibitem
\bibitem[Barducci \latin{et~al.}(2008)Barducci, Bussi, and
  Parrinello]{barducci2008well}
Barducci,~A.; Bussi,~G.; Parrinello,~M. Well-tempered metadynamics: a smoothly
  converging and tunable free-energy method. \emph{Physical review letters}
  \textbf{2008}, \emph{100}, 020603\relax
\mciteBstWouldAddEndPuncttrue
\mciteSetBstMidEndSepPunct{\mcitedefaultmidpunct}
{\mcitedefaultendpunct}{\mcitedefaultseppunct}\relax
\EndOfBibitem
\bibitem[Maragliano and Vanden-Eijnden(2006)Maragliano, and
  Vanden-Eijnden]{maragliano2006temperature}
Maragliano,~L.; Vanden-Eijnden,~E. A temperature accelerated method for
  sampling free energy and determining reaction pathways in rare events
  simulations. \emph{Chemical physics letters} \textbf{2006}, \emph{426},
  168--175\relax
\mciteBstWouldAddEndPuncttrue
\mciteSetBstMidEndSepPunct{\mcitedefaultmidpunct}
{\mcitedefaultendpunct}{\mcitedefaultseppunct}\relax
\EndOfBibitem
\bibitem[Abrams and Tuckerman(2008)Abrams, and Tuckerman]{abrams2008efficient}
Abrams,~J.~B.; Tuckerman,~M.~E. Efficient and direct generation of
  multidimensional free energy surfaces via adiabatic dynamics without
  coordinate transformations. \emph{The Journal of Physical Chemistry B}
  \textbf{2008}, \emph{112}, 15742--15757\relax
\mciteBstWouldAddEndPuncttrue
\mciteSetBstMidEndSepPunct{\mcitedefaultmidpunct}
{\mcitedefaultendpunct}{\mcitedefaultseppunct}\relax
\EndOfBibitem
\bibitem[Darve \latin{et~al.}(2008)Darve, Rodr{\'\i}guez-G{\'o}mez, and
  Pohorille]{darve2008adaptive}
Darve,~E.; Rodr{\'\i}guez-G{\'o}mez,~D.; Pohorille,~A. Adaptive biasing force
  method for scalar and vector free energy calculations. \emph{The Journal of
  chemical physics} \textbf{2008}, \emph{128}, 144120\relax
\mciteBstWouldAddEndPuncttrue
\mciteSetBstMidEndSepPunct{\mcitedefaultmidpunct}
{\mcitedefaultendpunct}{\mcitedefaultseppunct}\relax
\EndOfBibitem
\bibitem[Barducci \latin{et~al.}(2011)Barducci, Bonomi, and
  Parrinello]{barducci2011metadynamics}
Barducci,~A.; Bonomi,~M.; Parrinello,~M. Metadynamics. \emph{Wiley
  Interdisciplinary Reviews: Computational Molecular Science} \textbf{2011},
  \emph{1}, 826--843\relax
\mciteBstWouldAddEndPuncttrue
\mciteSetBstMidEndSepPunct{\mcitedefaultmidpunct}
{\mcitedefaultendpunct}{\mcitedefaultseppunct}\relax
\EndOfBibitem
\bibitem[Abrams and Vanden-Eijnden(2010)Abrams, and
  Vanden-Eijnden]{abrams2010large}
Abrams,~C.~F.; Vanden-Eijnden,~E. Large-scale conformational sampling of
  proteins using temperature-accelerated molecular dynamics. \emph{Biophysical
  Journal} \textbf{2010}, \emph{98}, 26a\relax
\mciteBstWouldAddEndPuncttrue
\mciteSetBstMidEndSepPunct{\mcitedefaultmidpunct}
{\mcitedefaultendpunct}{\mcitedefaultseppunct}\relax
\EndOfBibitem
\bibitem[Yu and Tuckerman(2011)Yu, and Tuckerman]{yu2011temperature}
Yu,~T.-Q.; Tuckerman,~M.~E. Temperature-accelerated method for exploring
  polymorphism in molecular crystals based on free energy. \emph{Physical
  review letters} \textbf{2011}, \emph{107}, 015701\relax
\mciteBstWouldAddEndPuncttrue
\mciteSetBstMidEndSepPunct{\mcitedefaultmidpunct}
{\mcitedefaultendpunct}{\mcitedefaultseppunct}\relax
\EndOfBibitem
\bibitem[Yu \latin{et~al.}(2014)Yu, Chen, Chen, Samanta, Vanden-Eijnden, and
  Tuckerman]{yu2014order}
Yu,~T.-Q.; Chen,~P.-Y.; Chen,~M.; Samanta,~A.; Vanden-Eijnden,~E.;
  Tuckerman,~M. Order-parameter-aided temperature-accelerated sampling for the
  exploration of crystal polymorphism and solid-liquid phase transitions.
  \emph{The Journal of chemical physics} \textbf{2014}, \emph{140},
  214109\relax
\mciteBstWouldAddEndPuncttrue
\mciteSetBstMidEndSepPunct{\mcitedefaultmidpunct}
{\mcitedefaultendpunct}{\mcitedefaultseppunct}\relax
\EndOfBibitem
\bibitem[Samanta \latin{et~al.}(2014)Samanta, Tuckerman, Yu, and
  E]{samanta2014microscopic}
Samanta,~A.; Tuckerman,~M.~E.; Yu,~T.-Q.; E,~W. Microscopic mechanisms of
  equilibrium melting of a solid. \emph{Science} \textbf{2014}, \emph{346},
  729--732\relax
\mciteBstWouldAddEndPuncttrue
\mciteSetBstMidEndSepPunct{\mcitedefaultmidpunct}
{\mcitedefaultendpunct}{\mcitedefaultseppunct}\relax
\EndOfBibitem
\bibitem[Tiwary and Parrinello(2013)Tiwary, and
  Parrinello]{tiwary2013metadynamics}
Tiwary,~P.; Parrinello,~M. From metadynamics to dynamics. \emph{Physical review
  letters} \textbf{2013}, \emph{111}, 230602\relax
\mciteBstWouldAddEndPuncttrue
\mciteSetBstMidEndSepPunct{\mcitedefaultmidpunct}
{\mcitedefaultendpunct}{\mcitedefaultseppunct}\relax
\EndOfBibitem
\bibitem[Fefferman \latin{et~al.}(2016)Fefferman, Mitter, and
  Narayanan]{fefferman2016testing}
Fefferman,~C.; Mitter,~S.; Narayanan,~H. Testing the manifold hypothesis.
  \emph{Journal of the American Mathematical Society} \textbf{2016}, \emph{29},
  983--1049\relax
\mciteBstWouldAddEndPuncttrue
\mciteSetBstMidEndSepPunct{\mcitedefaultmidpunct}
{\mcitedefaultendpunct}{\mcitedefaultseppunct}\relax
\EndOfBibitem
\bibitem[Narayanan and Mitter(2010)Narayanan, and Mitter]{narayanan2010sample}
Narayanan,~H.; Mitter,~S. Sample complexity of testing the manifold hypothesis.
  \emph{Advances in neural information processing systems} \textbf{2010},
  \emph{23}\relax
\mciteBstWouldAddEndPuncttrue
\mciteSetBstMidEndSepPunct{\mcitedefaultmidpunct}
{\mcitedefaultendpunct}{\mcitedefaultseppunct}\relax
\EndOfBibitem
\bibitem[Gorban and Tyukin(2018)Gorban, and Tyukin]{gorban2018blessing}
Gorban,~A.~N.; Tyukin,~I.~Y. Blessing of dimensionality: mathematical
  foundations of the statistical physics of data. \emph{Philosophical
  Transactions of the Royal Society A: Mathematical, Physical and Engineering
  Sciences} \textbf{2018}, \emph{376}, 20170237\relax
\mciteBstWouldAddEndPuncttrue
\mciteSetBstMidEndSepPunct{\mcitedefaultmidpunct}
{\mcitedefaultendpunct}{\mcitedefaultseppunct}\relax
\EndOfBibitem
\bibitem[Hummer and Szabo(2001)Hummer, and Szabo]{hummer2001free}
Hummer,~G.; Szabo,~A. Free energy reconstruction from nonequilibrium
  single-molecule pulling experiments. \emph{Proceedings of the National
  Academy of Sciences} \textbf{2001}, \emph{98}, 3658--3661\relax
\mciteBstWouldAddEndPuncttrue
\mciteSetBstMidEndSepPunct{\mcitedefaultmidpunct}
{\mcitedefaultendpunct}{\mcitedefaultseppunct}\relax
\EndOfBibitem
\bibitem[Bussi \latin{et~al.}(2006)Bussi, Gervasio, Laio, and
  Parrinello]{bussi2006free}
Bussi,~G.; Gervasio,~F.~L.; Laio,~A.; Parrinello,~M. Free-energy landscape for
  $\beta$ hairpin folding from combined parallel tempering and metadynamics.
  \emph{Journal of the American Chemical Society} \textbf{2006}, \emph{128},
  13435--13441\relax
\mciteBstWouldAddEndPuncttrue
\mciteSetBstMidEndSepPunct{\mcitedefaultmidpunct}
{\mcitedefaultendpunct}{\mcitedefaultseppunct}\relax
\EndOfBibitem
\bibitem[Granata \latin{et~al.}(2013)Granata, Camilloni, Vendruscolo, and
  Laio]{granata2013characterization}
Granata,~D.; Camilloni,~C.; Vendruscolo,~M.; Laio,~A. Characterization of the
  free-energy landscapes of proteins by NMR-guided metadynamics.
  \emph{Proceedings of the National Academy of Sciences} \textbf{2013},
  \emph{110}, 6817--6822\relax
\mciteBstWouldAddEndPuncttrue
\mciteSetBstMidEndSepPunct{\mcitedefaultmidpunct}
{\mcitedefaultendpunct}{\mcitedefaultseppunct}\relax
\EndOfBibitem
\bibitem[Das \latin{et~al.}(2006)Das, Moll, Stamati, Kavraki, and
  Clementi]{das2006low}
Das,~P.; Moll,~M.; Stamati,~H.; Kavraki,~L.~E.; Clementi,~C. Low-dimensional,
  free-energy landscapes of protein-folding reactions by nonlinear
  dimensionality reduction. \emph{Proceedings of the National Academy of
  Sciences} \textbf{2006}, \emph{103}, 9885--9890\relax
\mciteBstWouldAddEndPuncttrue
\mciteSetBstMidEndSepPunct{\mcitedefaultmidpunct}
{\mcitedefaultendpunct}{\mcitedefaultseppunct}\relax
\EndOfBibitem
\bibitem[Ferguson \latin{et~al.}(2011)Ferguson, Panagiotopoulos, Debenedetti,
  and Kevrekidis]{ferguson2011integrating}
Ferguson,~A.~L.; Panagiotopoulos,~A.~Z.; Debenedetti,~P.~G.; Kevrekidis,~I.~G.
  Integrating diffusion maps with umbrella sampling: Application to alanine
  dipeptide. \emph{The Journal of chemical physics} \textbf{2011}, \emph{134},
  135103\relax
\mciteBstWouldAddEndPuncttrue
\mciteSetBstMidEndSepPunct{\mcitedefaultmidpunct}
{\mcitedefaultendpunct}{\mcitedefaultseppunct}\relax
\EndOfBibitem
\bibitem[Rohrdanz \latin{et~al.}(2011)Rohrdanz, Zheng, Maggioni, and
  Clementi]{rohrdanz2011determination}
Rohrdanz,~M.~A.; Zheng,~W.; Maggioni,~M.; Clementi,~C. Determination of
  reaction coordinates via locally scaled diffusion map. \emph{The Journal of
  chemical physics} \textbf{2011}, \emph{134}, 124116\relax
\mciteBstWouldAddEndPuncttrue
\mciteSetBstMidEndSepPunct{\mcitedefaultmidpunct}
{\mcitedefaultendpunct}{\mcitedefaultseppunct}\relax
\EndOfBibitem
\bibitem[No{\'e} and Clementi(2015)No{\'e}, and Clementi]{noe2015kinetic}
No{\'e},~F.; Clementi,~C. Kinetic distance and kinetic maps from molecular
  dynamics simulation. \emph{Journal of chemical theory and computation}
  \textbf{2015}, \emph{11}, 5002--5011\relax
\mciteBstWouldAddEndPuncttrue
\mciteSetBstMidEndSepPunct{\mcitedefaultmidpunct}
{\mcitedefaultendpunct}{\mcitedefaultseppunct}\relax
\EndOfBibitem
\bibitem[Zhang and Chen(2018)Zhang, and Chen]{zhang2018unfolding}
Zhang,~J.; Chen,~M. Unfolding hidden barriers by active enhanced sampling.
  \emph{Physical review letters} \textbf{2018}, \emph{121}, 010601\relax
\mciteBstWouldAddEndPuncttrue
\mciteSetBstMidEndSepPunct{\mcitedefaultmidpunct}
{\mcitedefaultendpunct}{\mcitedefaultseppunct}\relax
\EndOfBibitem
\bibitem[Fu \latin{et~al.}(2022)Fu, Xie, Rebello, Olsen, and
  Jaakkola]{fu2022simulate}
Fu,~X.; Xie,~T.; Rebello,~N.~J.; Olsen,~B.~D.; Jaakkola,~T. Simulate
  time-integrated coarse-grained molecular dynamics with geometric machine
  learning. \emph{arXiv preprint arXiv:2204.10348} \textbf{2022}, \relax
\mciteBstWouldAddEndPunctfalse
\mciteSetBstMidEndSepPunct{\mcitedefaultmidpunct}
{}{\mcitedefaultseppunct}\relax
\EndOfBibitem
\bibitem[Ceriotti \latin{et~al.}(2011)Ceriotti, Tribello, and
  Parrinello]{ceriotti2011simplifying}
Ceriotti,~M.; Tribello,~G.~A.; Parrinello,~M. Simplifying the representation of
  complex free-energy landscapes using sketch-map. \emph{Proceedings of the
  National Academy of Sciences} \textbf{2011}, \emph{108}, 13023--13028\relax
\mciteBstWouldAddEndPuncttrue
\mciteSetBstMidEndSepPunct{\mcitedefaultmidpunct}
{\mcitedefaultendpunct}{\mcitedefaultseppunct}\relax
\EndOfBibitem
\bibitem[Tiwary and Berne(2016)Tiwary, and Berne]{tiwary2016spectral}
Tiwary,~P.; Berne,~B. Spectral gap optimization of order parameters for
  sampling complex molecular systems. \emph{Proceedings of the National Academy
  of Sciences} \textbf{2016}, \emph{113}, 2839--2844\relax
\mciteBstWouldAddEndPuncttrue
\mciteSetBstMidEndSepPunct{\mcitedefaultmidpunct}
{\mcitedefaultendpunct}{\mcitedefaultseppunct}\relax
\EndOfBibitem
\bibitem[Mendels \latin{et~al.}(2018)Mendels, Piccini, and
  Parrinello]{mendels2018collective}
Mendels,~D.; Piccini,~G.; Parrinello,~M. Collective variables from local
  fluctuations. \emph{The journal of physical chemistry letters} \textbf{2018},
  \emph{9}, 2776--2781\relax
\mciteBstWouldAddEndPuncttrue
\mciteSetBstMidEndSepPunct{\mcitedefaultmidpunct}
{\mcitedefaultendpunct}{\mcitedefaultseppunct}\relax
\EndOfBibitem
\bibitem[Ribeiro \latin{et~al.}(2018)Ribeiro, Bravo, Wang, and
  Tiwary]{ribeiro2018reweighted}
Ribeiro,~J. M.~L.; Bravo,~P.; Wang,~Y.; Tiwary,~P. Reweighted autoencoded
  variational Bayes for enhanced sampling (RAVE). \emph{The Journal of chemical
  physics} \textbf{2018}, \emph{149}, 072301\relax
\mciteBstWouldAddEndPuncttrue
\mciteSetBstMidEndSepPunct{\mcitedefaultmidpunct}
{\mcitedefaultendpunct}{\mcitedefaultseppunct}\relax
\EndOfBibitem
\bibitem[Wang \latin{et~al.}(2019)Wang, Ribeiro, and Tiwary]{wang2019past}
Wang,~Y.; Ribeiro,~J. M.~L.; Tiwary,~P. Past--future information bottleneck for
  sampling molecular reaction coordinate simultaneously with thermodynamics and
  kinetics. \emph{Nature communications} \textbf{2019}, \emph{10}, 3573\relax
\mciteBstWouldAddEndPuncttrue
\mciteSetBstMidEndSepPunct{\mcitedefaultmidpunct}
{\mcitedefaultendpunct}{\mcitedefaultseppunct}\relax
\EndOfBibitem
\bibitem[Nuske \latin{et~al.}(2014)Nuske, Keller, P{\'e}rez-Hern{\'a}ndez, Mey,
  and No{\'e}]{nuske2014variational}
Nuske,~F.; Keller,~B.~G.; P{\'e}rez-Hern{\'a}ndez,~G.; Mey,~A.~S.; No{\'e},~F.
  Variational approach to molecular kinetics. \emph{Journal of chemical theory
  and computation} \textbf{2014}, \emph{10}, 1739--1752\relax
\mciteBstWouldAddEndPuncttrue
\mciteSetBstMidEndSepPunct{\mcitedefaultmidpunct}
{\mcitedefaultendpunct}{\mcitedefaultseppunct}\relax
\EndOfBibitem
\bibitem[Facco \latin{et~al.}(2017)Facco, d’Errico, Rodriguez, and
  Laio]{facco2017estimating}
Facco,~E.; d’Errico,~M.; Rodriguez,~A.; Laio,~A. Estimating the intrinsic
  dimension of datasets by a minimal neighborhood information. \emph{Scientific
  reports} \textbf{2017}, \emph{7}, 12140\relax
\mciteBstWouldAddEndPuncttrue
\mciteSetBstMidEndSepPunct{\mcitedefaultmidpunct}
{\mcitedefaultendpunct}{\mcitedefaultseppunct}\relax
\EndOfBibitem
\bibitem[Findeis(2007)]{findeis2007role}
Findeis,~M.~A. The role of amyloid $\beta$ peptide 42 in Alzheimer's disease.
  \emph{Pharmacology \& therapeutics} \textbf{2007}, \emph{116}, 266--286\relax
\mciteBstWouldAddEndPuncttrue
\mciteSetBstMidEndSepPunct{\mcitedefaultmidpunct}
{\mcitedefaultendpunct}{\mcitedefaultseppunct}\relax
\EndOfBibitem
\bibitem[Chen \latin{et~al.}(2012)Chen, Cuendet, and
  Tuckerman]{chen2012heating}
Chen,~M.; Cuendet,~M.~A.; Tuckerman,~M.~E. Heating and flooding: A unified
  approach for rapid generation of free energy surfaces. \emph{The Journal of
  Chemical Physics} \textbf{2012}, \emph{137}, 024102\relax
\mciteBstWouldAddEndPuncttrue
\mciteSetBstMidEndSepPunct{\mcitedefaultmidpunct}
{\mcitedefaultendpunct}{\mcitedefaultseppunct}\relax
\EndOfBibitem
\bibitem[Debnath and Parrinello(2020)Debnath, and Parrinello]{Parrinello2020gm}
Debnath,~J.; Parrinello,~M. Gaussian Mixture-Based Enhanced Sampling for
  Statics and Dynamics. \emph{The Journal of Physical Chemistry Letters}
  \textbf{2020}, \emph{11}, 5076--5080, PMID: 32510225\relax
\mciteBstWouldAddEndPuncttrue
\mciteSetBstMidEndSepPunct{\mcitedefaultmidpunct}
{\mcitedefaultendpunct}{\mcitedefaultseppunct}\relax
\EndOfBibitem
\bibitem[Zhang \latin{et~al.}(2018)Zhang, Wang, and E]{E2018reinforced}
Zhang,~L.; Wang,~H.; E,~W. {Reinforced dynamics for enhanced sampling in large
  atomic and molecular systems}. \emph{The Journal of Chemical Physics}
  \textbf{2018}, \emph{148}, 124113\relax
\mciteBstWouldAddEndPuncttrue
\mciteSetBstMidEndSepPunct{\mcitedefaultmidpunct}
{\mcitedefaultendpunct}{\mcitedefaultseppunct}\relax
\EndOfBibitem
\bibitem[Zhang \latin{et~al.}(2020)Zhang, Lei, Yang, and Gao]{Gao2020vade}
Zhang,~J.; Lei,~Y.-K.; Yang,~Y.~I.; Gao,~Y.~Q. {Deep learning for variational
  multiscale molecular modeling}. \emph{The Journal of Chemical Physics}
  \textbf{2020}, \emph{153}, 174115\relax
\mciteBstWouldAddEndPuncttrue
\mciteSetBstMidEndSepPunct{\mcitedefaultmidpunct}
{\mcitedefaultendpunct}{\mcitedefaultseppunct}\relax
\EndOfBibitem
\bibitem[Cuendet and Tuckerman(2014)Cuendet, and Tuckerman]{cuendet2014free}
Cuendet,~M.~A.; Tuckerman,~M.~E. Free energy reconstruction from metadynamics
  or adiabatic free energy dynamics simulations. \emph{Journal of chemical
  theory and computation} \textbf{2014}, \emph{10}, 2975--2986\relax
\mciteBstWouldAddEndPuncttrue
\mciteSetBstMidEndSepPunct{\mcitedefaultmidpunct}
{\mcitedefaultendpunct}{\mcitedefaultseppunct}\relax
\EndOfBibitem
\bibitem[Chen(2017)]{chen2017tutorial}
Chen,~Y.-C. A tutorial on kernel density estimation and recent advances.
  \emph{Biostatistics \& Epidemiology} \textbf{2017}, \emph{1}, 161--187\relax
\mciteBstWouldAddEndPuncttrue
\mciteSetBstMidEndSepPunct{\mcitedefaultmidpunct}
{\mcitedefaultendpunct}{\mcitedefaultseppunct}\relax
\EndOfBibitem
\bibitem[Mack and Rosenblatt(1979)Mack, and Rosenblatt]{mack1979multivariate}
Mack,~Y.; Rosenblatt,~M. Multivariate k-nearest neighbor density estimates.
  \emph{Journal of Multivariate Analysis} \textbf{1979}, \emph{9}, 1--15\relax
\mciteBstWouldAddEndPuncttrue
\mciteSetBstMidEndSepPunct{\mcitedefaultmidpunct}
{\mcitedefaultendpunct}{\mcitedefaultseppunct}\relax
\EndOfBibitem
\bibitem[Reynolds \latin{et~al.}(2009)Reynolds, \latin{et~al.}
  others]{reynolds2009gaussian}
Reynolds,~D.~A., \latin{et~al.}  Gaussian mixture models. \emph{Encyclopedia of
  biometrics} \textbf{2009}, \emph{741}\relax
\mciteBstWouldAddEndPuncttrue
\mciteSetBstMidEndSepPunct{\mcitedefaultmidpunct}
{\mcitedefaultendpunct}{\mcitedefaultseppunct}\relax
\EndOfBibitem
\bibitem[Song \latin{et~al.}(2021)Song, Sohl-Dickstein, Kingma, Kumar, Ermon,
  and Poole]{song2021score}
Song,~Y.; Sohl-Dickstein,~J.; Kingma,~D.~P.; Kumar,~A.; Ermon,~S.; Poole,~B.
  Score-Based Generative Modeling through Stochastic Differential Equations.
  International Conference on Learning Representations. 2021\relax
\mciteBstWouldAddEndPuncttrue
\mciteSetBstMidEndSepPunct{\mcitedefaultmidpunct}
{\mcitedefaultendpunct}{\mcitedefaultseppunct}\relax
\EndOfBibitem
\bibitem[De~Bortoli \latin{et~al.}(2022)De~Bortoli, Mathieu, Hutchinson,
  Thornton, Teh, and Doucet]{de2022riemannian}
De~Bortoli,~V.; Mathieu,~E.; Hutchinson,~M.; Thornton,~J.; Teh,~Y.~W.;
  Doucet,~A. Riemannian score-based generative modelling. \emph{Advances in
  Neural Information Processing Systems} \textbf{2022}, \emph{35},
  2406--2422\relax
\mciteBstWouldAddEndPuncttrue
\mciteSetBstMidEndSepPunct{\mcitedefaultmidpunct}
{\mcitedefaultendpunct}{\mcitedefaultseppunct}\relax
\EndOfBibitem
\bibitem[Anderson(1982)]{anderson1982reverse}
Anderson,~B.~D. Reverse-time diffusion equation models. \emph{Stochastic
  Processes and their Applications} \textbf{1982}, \emph{12}, 313--326\relax
\mciteBstWouldAddEndPuncttrue
\mciteSetBstMidEndSepPunct{\mcitedefaultmidpunct}
{\mcitedefaultendpunct}{\mcitedefaultseppunct}\relax
\EndOfBibitem
\bibitem[He \latin{et~al.}(2016)He, Zhang, Ren, and Sun]{he2016deep}
He,~K.; Zhang,~X.; Ren,~S.; Sun,~J. Deep residual learning for image
  recognition. Proceedings of the IEEE conference on computer vision and
  pattern recognition. 2016; pp 770--778\relax
\mciteBstWouldAddEndPuncttrue
\mciteSetBstMidEndSepPunct{\mcitedefaultmidpunct}
{\mcitedefaultendpunct}{\mcitedefaultseppunct}\relax
\EndOfBibitem
\bibitem[Ronneberger \latin{et~al.}(2015)Ronneberger, Fischer, and
  Brox]{olaf2015unet}
Ronneberger,~O.; Fischer,~P.; Brox,~T. U-Net: Convolutional Networks for
  Biomedical Image Segmentation. Medical Image Computing and Computer-Assisted
  Intervention -- MICCAI 2015. Cham, 2015; pp 234--241\relax
\mciteBstWouldAddEndPuncttrue
\mciteSetBstMidEndSepPunct{\mcitedefaultmidpunct}
{\mcitedefaultendpunct}{\mcitedefaultseppunct}\relax
\EndOfBibitem
\bibitem[Scarselli \latin{et~al.}(2008)Scarselli, Gori, Tsoi, Hagenbuchner, and
  Monfardini]{scarselli2008graph}
Scarselli,~F.; Gori,~M.; Tsoi,~A.~C.; Hagenbuchner,~M.; Monfardini,~G. The
  graph neural network model. \emph{IEEE transactions on neural networks}
  \textbf{2008}, \emph{20}, 61--80\relax
\mciteBstWouldAddEndPuncttrue
\mciteSetBstMidEndSepPunct{\mcitedefaultmidpunct}
{\mcitedefaultendpunct}{\mcitedefaultseppunct}\relax
\EndOfBibitem
\bibitem[Rezende \latin{et~al.}(2020)Rezende, Papamakarios, Racaniere, Albergo,
  Kanwar, Shanahan, and Cranmer]{rezende2020normalizing}
Rezende,~D.~J.; Papamakarios,~G.; Racaniere,~S.; Albergo,~M.; Kanwar,~G.;
  Shanahan,~P.; Cranmer,~K. Normalizing flows on tori and spheres.
  International Conference on Machine Learning. 2020; pp 8083--8092\relax
\mciteBstWouldAddEndPuncttrue
\mciteSetBstMidEndSepPunct{\mcitedefaultmidpunct}
{\mcitedefaultendpunct}{\mcitedefaultseppunct}\relax
\EndOfBibitem
\bibitem[K{\"o}hler \latin{et~al.}(2021)K{\"o}hler, Kr{\"a}mer, and
  No{\'e}]{kohler2021smooth}
K{\"o}hler,~J.; Kr{\"a}mer,~A.; No{\'e},~F. Smooth normalizing flows.
  \emph{Advances in Neural Information Processing Systems} \textbf{2021},
  \emph{34}, 2796--2809\relax
\mciteBstWouldAddEndPuncttrue
\mciteSetBstMidEndSepPunct{\mcitedefaultmidpunct}
{\mcitedefaultendpunct}{\mcitedefaultseppunct}\relax
\EndOfBibitem
\bibitem[Strodel and Wales(2008)Strodel, and Wales]{strodel2008free}
Strodel,~B.; Wales,~D.~J. Free energy surfaces from an extended harmonic
  superposition approach and kinetics for alanine dipeptide. \emph{Chemical
  Physics Letters} \textbf{2008}, \emph{466}, 105--115\relax
\mciteBstWouldAddEndPuncttrue
\mciteSetBstMidEndSepPunct{\mcitedefaultmidpunct}
{\mcitedefaultendpunct}{\mcitedefaultseppunct}\relax
\EndOfBibitem
\bibitem[Smith(1999)]{smith1999alanine}
Smith,~P.~E. The alanine dipeptide free energy surface in solution. \emph{The
  Journal of chemical physics} \textbf{1999}, \emph{111}, 5568--5579\relax
\mciteBstWouldAddEndPuncttrue
\mciteSetBstMidEndSepPunct{\mcitedefaultmidpunct}
{\mcitedefaultendpunct}{\mcitedefaultseppunct}\relax
\EndOfBibitem
\bibitem[Hénin \latin{et~al.}(2010)Hénin, Fiorin, Chipot, and
  Klein]{Klein2010tdbias}
Hénin,~J.; Fiorin,~G.; Chipot,~C.; Klein,~M.~L. Exploring Multidimensional
  Free Energy Landscapes Using Time-Dependent Biases on Collective Variables.
  \emph{Journal of Chemical Theory and Computation} \textbf{2010}, \emph{6},
  35--47, PMID: 26614317\relax
\mciteBstWouldAddEndPuncttrue
\mciteSetBstMidEndSepPunct{\mcitedefaultmidpunct}
{\mcitedefaultendpunct}{\mcitedefaultseppunct}\relax
\EndOfBibitem
\bibitem[Chen \latin{et~al.}(2015)Chen, Yu, and Tuckerman]{chen2015locating}
Chen,~M.; Yu,~T.-Q.; Tuckerman,~M.~E. Locating landmarks on high-dimensional
  free energy surfaces. \emph{Proceedings of the National Academy of Sciences}
  \textbf{2015}, \emph{112}, 3235--3240\relax
\mciteBstWouldAddEndPuncttrue
\mciteSetBstMidEndSepPunct{\mcitedefaultmidpunct}
{\mcitedefaultendpunct}{\mcitedefaultseppunct}\relax
\EndOfBibitem
\bibitem[Sutto \latin{et~al.}(2010)Sutto, D’Abramo, and
  Gervasio]{sutto2010comparing}
Sutto,~L.; D’Abramo,~M.; Gervasio,~F.~L. Comparing the efficiency of biased
  and unbiased molecular dynamics in reconstructing the free energy landscape
  of met-enkephalin. \emph{Journal of Chemical Theory and Computation}
  \textbf{2010}, \emph{6}, 3640--3646\relax
\mciteBstWouldAddEndPuncttrue
\mciteSetBstMidEndSepPunct{\mcitedefaultmidpunct}
{\mcitedefaultendpunct}{\mcitedefaultseppunct}\relax
\EndOfBibitem
\bibitem[Sutto \latin{et~al.}(2012)Sutto, Marsili, and Gervasio]{sutto2012new}
Sutto,~L.; Marsili,~S.; Gervasio,~F.~L. New advances in metadynamics.
  \emph{Wiley Interdisciplinary Reviews: Computational Molecular Science}
  \textbf{2012}, \emph{2}, 771--779\relax
\mciteBstWouldAddEndPuncttrue
\mciteSetBstMidEndSepPunct{\mcitedefaultmidpunct}
{\mcitedefaultendpunct}{\mcitedefaultseppunct}\relax
\EndOfBibitem
\bibitem[Sicard and Senet(2013)Sicard, and Senet]{sicard2013reconstructing}
Sicard,~F.; Senet,~P. Reconstructing the free-energy landscape of
  Met-enkephalin using dihedral principal component analysis and well-tempered
  metadynamics. \emph{The Journal of Chemical Physics} \textbf{2013},
  \emph{138}, 235101\relax
\mciteBstWouldAddEndPuncttrue
\mciteSetBstMidEndSepPunct{\mcitedefaultmidpunct}
{\mcitedefaultendpunct}{\mcitedefaultseppunct}\relax
\EndOfBibitem
\end{mcitethebibliography}
\end{document}